\documentclass{vgtc}                          % final (conference style)
\ifpdf%                                % if we use pdflatex
  \pdfoutput=1\relax                   % create PDFs from pdfLaTeX
  \usepackage{graphicx}                % allow us to embed graphics files
  \DeclareGraphicsExtensions{.pdf,.png,.jpg,.jpeg} % for pdflatex we expect .pdf, .png, or .jpg files
\else%                                 % else we use pure latex
  \usepackage{graphicx}                % allow us to embed graphics files
  \DeclareGraphicsExtensions{.eps}     % for pure latex we expect eps files
\fi%

\graphicspath{{figures/}{pictures/}{images/}{./}} % where to search for the images
\usepackage{booktabs}
\usepackage{amsmath}
\usepackage{microtype}                 % use micro-typography (slightly more compact, better to read)
\PassOptionsToPackage{warn}{textcomp}  % to address font issues with \textrightarrow
\usepackage{textcomp}                  % use better special symbols
\usepackage{mathptmx}                  % use matching math font
\usepackage{times}                     % we use Times as the main font
\usepackage{cite}                      % needed to automatically sort the references
\usepackage{tabu}                      % only used for the table example
\usepackage{booktabs}                  % only used for the table example
\usepackage{pgfplots}
\usepackage{comment}
\pgfplotsset{compat=1.17}

\vgtccategory{Research}

\title{Evaluating CrowdSplat: Perceived Level of Detail for Gaussian Crowds}

\author{Xiaohan Sun\thanks{These authors contributed equally to this work.\\
\hspace*{1.7em}email: \texttt{\{sunx4|yixu\}@tcd.ie}} %
\and Yinghan Xu\footnotemark[1] %
\and John Dingliana %
\and Carol O'Sullivan}
\affiliation{\scriptsize Trinity College Dublin, Ireland}

\teaser{
  \centering  \includegraphics[width=0.9\linewidth]{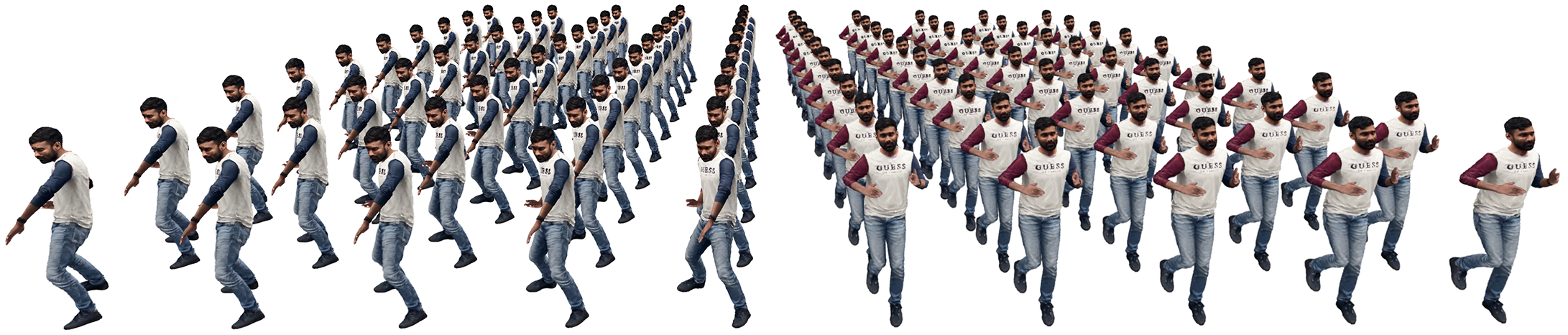}
  \caption{3D rendering of a crowd with four levels of detail (202k, 50k, 12k, or 3k Gaussians), chosen based on our results.}
  \label{fig:teaser}
}

\abstract{Efficient and realistic crowd rendering is an important element of many real-time graphics applications such as Virtual Reality (VR) and games. To this end, Levels of Detail (LOD) avatar representations such as polygonal meshes, image-based impostors, and point clouds have been proposed and evaluated. More recently, 3D Gaussian Splatting has been explored as a potential method for real-time crowd rendering.  In this paper, we present a two-alternative forced choice (2AFC) experiment that aims to determine the perceived quality of 3D Gaussian avatars. Three factors were explored: \emph{Motion}, \emph{LOD} (i.e., \#Gaussians), and the avatar height in \emph{Pixels} (corresponding to the viewing distance). Participants viewed pairs of animated 3D Gaussian avatars and were tasked with choosing the most detailed one. Our findings can inform the optimization of LOD strategies in Gaussian-based crowd rendering, thereby helping to achieve efficient rendering while maintaining visual quality in real-time applications.%
} % end of abstract

\CCScatlist{
   \CCScatTwelve{Computing methodologies}{Computer graphics}{Rendering}{Perception; Crowd rendering}
}

\begin{document}
\maketitle

%\firstsection{Introduction}
%\vspace{-0.2cm}
\section{Introduction}
\label{Intro}
The simulation of realistic crowd scenarios has been an active area of research for many decades. Lemonari et al. \cite{Lemonari:2022:Authoring} provide a comprehensive survey of the various components of this challenge, which include high-level behaviors, motion planning, animation and visualization. In this paper, we focus on the problem of 3D Gaussian Splatting for crowd rendering \cite{CrowdSplat}, and evaluate how several factors may affect how viewers perceive these Gaussian crowds. 

Adaptive LOD strategies for Gaussian avatars~\cite{RW4} have shown promise in balancing computational costs with visual fidelity, particularly for virtual environments. However, although the perception of LOD human representations has previously been studied (e.g., \cite{RW2}), the perceptual impact of using real-image-based Gaussian avatars for crowd rendering remains under-explored. We present a user study in which we aim to answer the following questions about Gaussian avatars:
\vspace{-0.1cm}
\begin{enumerate}
    \item[Q1.] Are rendering artefacts more noticeable for simpler cyclical \emph{Motions} than for more complex ones?
    \vspace{-0.2cm}
    \item[Q2.] Does reducing the \emph{LOD} (i.e., number of Gaussians) result in higher visibility of artefacts?
     \vspace{-0.2cm}
    \item[Q3.] Are artefacts less noticeable if the avatar covers a smaller number of the image \emph{Pixels} (i.e., increasing the viewing distance)?
     \vspace{-0.5cm}
    \item[Q4.] Will the noticeability of artefacts vary based on different combinations of Motion, LOD and Pixels?
     \vspace{-0.2cm}
\end{enumerate}

In a two-alternative forced choice (2AFC) experiment (Section \ref{sec:experiment}), we studied the effects of Motion, LOD and Pixels and their interactions on viewer perception of Gaussian avatars. Participants viewed pairs of animated
3D Gaussian avatars, where one was rendered at the highest quality (gold standard), while the other one was rendered at a lower LOD. One group of participants viewed avatars performing a cyclical jogging motion, while the other viewed them more complex martial arts performance. In each case, the task was to choose the most detailed of the two avatars shown. A three-way Analysis of Variance (ANOVA) with independent variables Motion (2) $\times$ LOD (3) $\times$ Pixels (5) and independent variable Accuracy (i.e., the proportion of times they correctly chose the gold standard) was conducted, followed by Bonferroni post-hoc tests on significant effects. Higher accuracy therefore indicates that visual artefacts were more noticeable.

We found that, for Q1, the answer is ``No''. There was a main effect of Motion, and higher accuracy was observed for the complex motion than for the cyclical one, indicating that the artefacts were more noticeable for the former. For Q2,  the answer is ``Yes'', as accuracy (and concomitantly, detection of artefacts) decreased with increasing numbers of Gaussians. The answer to Q3 is ``Not always'', as no main effect of Pixels was found. However, to answer Q4, Pixels did interact with LOD, where viewers were most accurate when the avatar was rendered at the lowest LOD and with the lowest number of pixels (i.e., at the furthest viewing distance). No other significant interaction effects were found. 

Our results were used to guide the placement of Gaussian crowd avatars in a real-time animated crowd scene (see Figure \ref{fig:teaser}). Insights gained can help to optimize LOD strategies and achieve efficient
rendering while maintaining visual quality in real-time applications.
\vspace{-0.6cm}
\section{Related Work}
\vspace{-0.1cm}
Level of Detail (LOD) techniques are used to efficiently render large-scale crowds by reducing the visual complexity of avatars based on their screen contribution. The key idea is that avatars further from the camera require less detail, thus allowing computational resources to be focused on foreground elements, where visual fidelity is most critical. We review the various avatar LOD representations that have been used previously, and then discuss the use of 3D Gaussian Splatting (3DGS)~\cite{TBP08} as an alternative to these approaches.

\begin{figure*}[t]
   \centering    
    \includegraphics[width=0.9\linewidth]{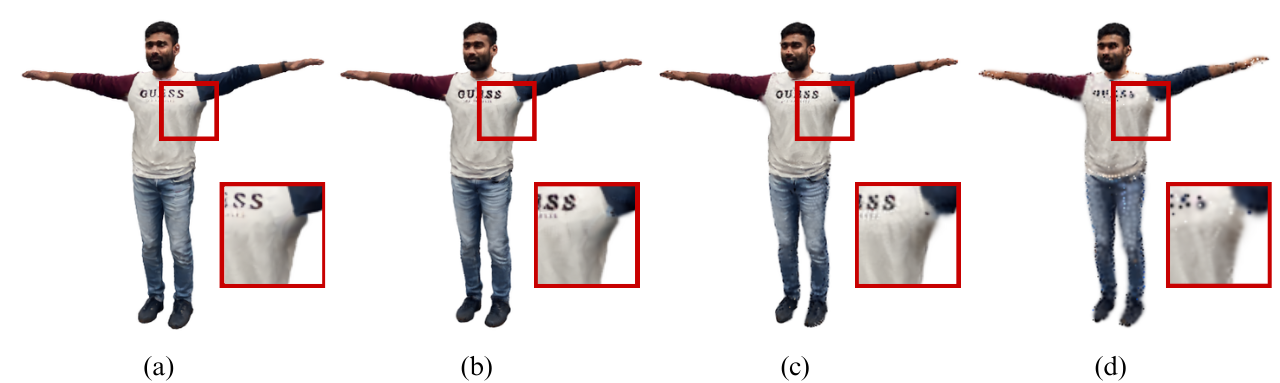}
    \caption{The Gaussian avatar at different LOD (\#Gaussians):
    %at 3.0\,m. 
    (a) Gold Standard--202k, 
    (b) LOD3--50k, 
    (c) LOD2--12k, 
    (d) LOD1--3k.}
    \label{fig:LOD}
\end{figure*}

\noindent
\textbf{LOD Representations:} One common representation of LOD avatars is to model them as detailed \emph{polygonal meshes}, which excel at delivering geometric fidelity. Technical solutions such as tessellation shaders allow for dynamically enhancement of low detailed meshes, thereby enabling close-up realism without excessive memory usage~\cite{TBP08}. However, simplifying animated meshes often introduces errors in attributes like skinning weights, leading to visible artefacts~\cite{LS09}. 

As an alternative, \emph{point-based} techniques use collections of points to depict avatars, offering computational efficiency for distant scenes~\cite{Bar05}. While efficient, these methods frequently result in artefacts when viewed up close, which limits their applicability for scenes requiring high detail. Hybrid approaches that integrate point and polygon representations~\cite{WS02} improve consistency across varying distances.

\emph{Image-based} representations offer an alternative solution by replacing complex 3D geometry with precomputed 2D impostors, which significantly reduce rendering costs~\cite{I1,TLC2002}. Such methods are particularly effective for background avatars but struggle with dynamic motions and close-up perspectives. Hybrid approaches have also been proposed to address these problems. Dobbyn et al.~\cite{I3} presented \emph{Geopostors}, where 3D geometry is used for foreground characters and image-based impostors for the background. Shaders are used to efficiently enhance the realism and variety of the dynamically-lit impostors, thereby allowing the two representations to be switch interchangeably based on the ``pixel to texel'' ratio of each avatar's current distance from the viewer. Popping artefacts are thus reduced and the switches are almost imperceptible. However, the memory requirements for multiple frames of animation rendered from different viewpoints for each impostor reduces the practicality of this approach. Kavan et al.~\cite{PP} aimed to address this limitation by developing \emph{Polypostors}, which augments the Geopostor approach by incorporating skeletal articulation, thereby balancing rendering efficiency, visual quality and memory requirements. Based on this idea, Beacco et al. ~\cite{Beacco11, Beacco12} rendered each joint with either relief or flat impostors, thus allowing for real time animation without increasing memory requirements %These advancements reflect ongoing efforts to refine LOD strategies across different scenarios.

In prior work, the \emph{perception} of LOD avatars in crowd rendering has also been explored. For instance, studies have examined how crowd density and viewpoint influence the perception of group behaviors, offering guidance on when such behaviors are necessary for realism~\cite{RW1}. Other work has compared the effectiveness of simplified meshes and impostors in conveying motion and appearance, identifying perceptual thresholds for different representations~\cite{RW2}. Additionally, perceptual evaluations of deformable clothing in LOD models have provided insights into optimizing representations for realistically clothed avatars~\cite{RW3}. The effect of color, texture and motion variety of LOD avatars on perception has also been explored by McDonnell et al.~\cite{mcdonnell2008clone}.

%\subsection{3D Gaussian Splatting for Animated Crowds}
\noindent
\textbf{3D Gaussian Splatting for Animated Crowds: }
Recent advancements in 3D Gaussian Splatting (3DGS) have enabled efficient and high-quality rendering for complex visual scenes \cite{3dgs}. This technique represents objects as anisotropic 3D Gaussian primitives, effectively blending light and color in a computationally efficient manner. Several researchers have applied 3DGS to create avatars for real-time photorealistic rendering. While Qian et al.~\cite{qian20233dgsavatar} and Zielonka et al.~\cite{zielonka25dega} use multi-view videos to reconstruct and animate realistic humans, others  achieve realistic avatars with dynamic 3D appearances using only a single video \cite{hu2023gauhuman, hugs, GSA}. The GaussianAvatar\cite{GSA} method results in better image quality than those that use a NERF-based method for 3D human reconstruction from a single video \cite{Jiang2022InstantAvatarLA, Weng2022HumanNeRFFR},  Additionally, Liu et al.~\cite{liu2024citygaussian} and Kerbl et al.~\cite{hierarchicalgaussians24} have demonstrated the technique's scalability and real-time rendering capabilities for large, static scenes.
\vspace{0.5cm}

While NERF-based methods have not been used to date for crowd rendering, 3D Gaussian Splatting offers advantages over traditional LOD representations in terms of realistic reconstruction and rendering. Sun et al.~\cite{CrowdSplat} recently presented CrowdSplat, which uses common 3D Gaussian attributes from an avatar template to optimize memory usage and rendering speed for large animated crowds. Although this work demonstrated how Gaussian-based LOD strategies could efficiently render thousands of avatars with high visual fidelity, there was no systematic evaluation of how viewers perceive these technical optimizations. In this paper, we build upon the CrowdSplat framework to evaluate the perception of 3DGS avatars under different conditions. In a user study, we examine when viewers notice differences in the 3D Gaussian count at various distances and how different types of motion affect perception. We offer practical guidelines for optimizing Gaussian-based LOD strategies in real-time crowd rendering, thus striking a balance between computational efficiency and perceived visual quality.

\begin{figure}[t]
    \centering    \includegraphics[width=0.8\linewidth]{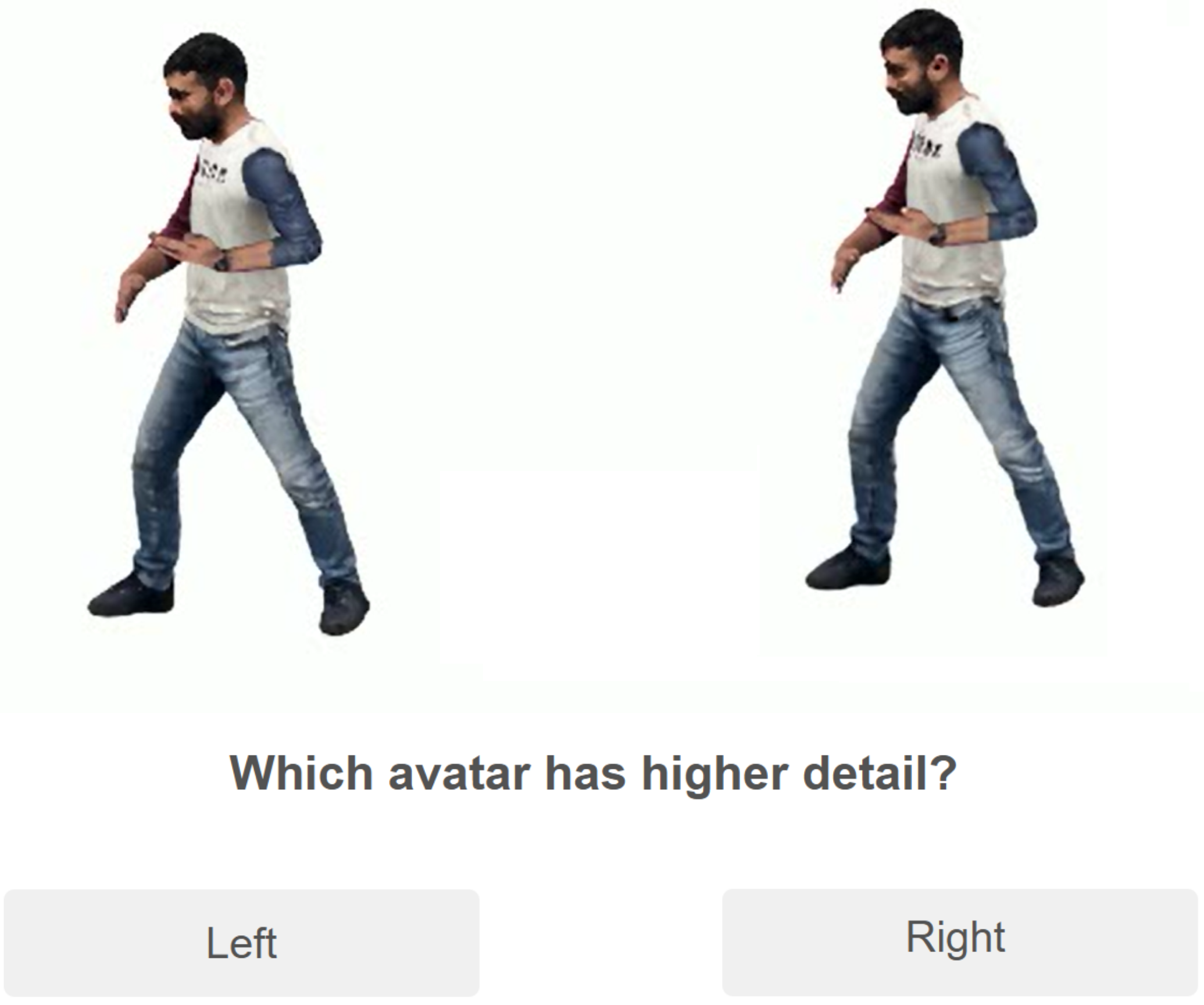}
    \caption{An example of the video stimuli.}
    \label{fig:design}
\end{figure}

\section{Experiment Design}
\label{sec:experiment}
%\subsection{Dataset and Setup}
\textbf{Stimuli: }
In our experiment, we aimed to evaluate the perceived LOD of 3DGS avatars by systematically varying the number of reconstructed 3D Gaussians and viewing distances. Following the approach of GaussianAvatar~\cite{GSA}, we utilized monocular video recordings of a subject performing random movements for two minutes as training data. To reconstruct the avatar templates, we generated positional maps at different resolutions in $64\times64$, $128\times128$, $256\times256$ and $512\times512$, resulting in four levels of detail (LOD) represented by different number of 3D Gaussians: 3k, 12k, 50k and 202k, the latter being the gold standard (GS), as shown in Figure~\ref{fig:LOD}. An example stimulus, showing the GS and a lower LOD, can be seen in Figure~\ref{fig:design}.

\begin{figure*}[t]
    \centering
    \includegraphics[width=0.9\linewidth]{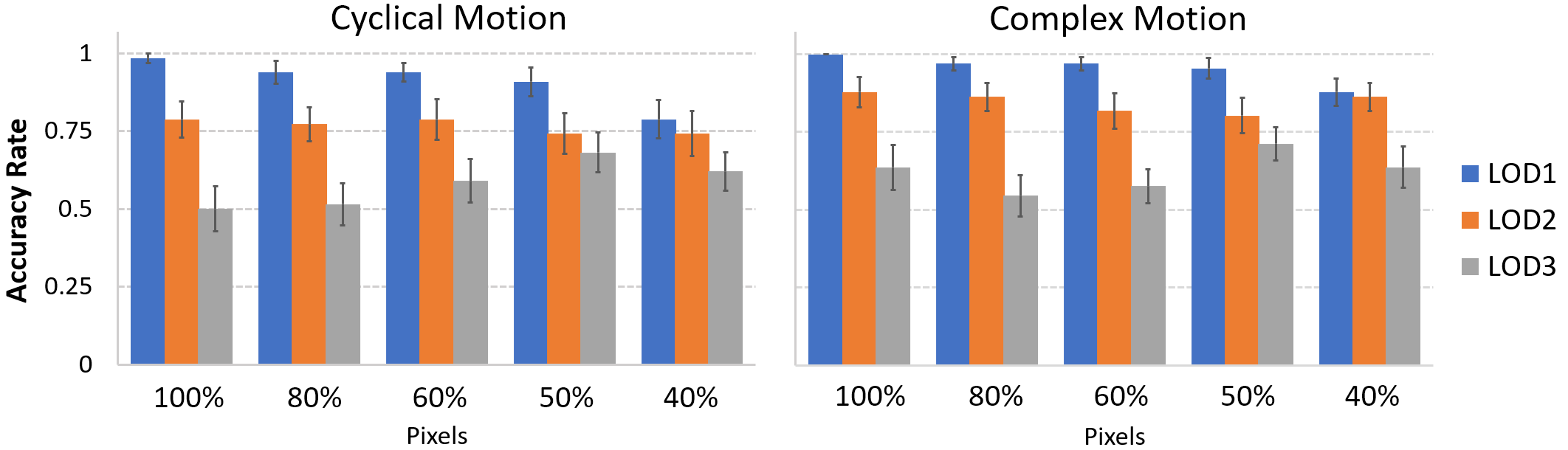}
    \caption{Accuracy rates by viewing distance/pixels and LOD for cyclical and complex motions.}
    \label{fig:expresults}
\end{figure*}

\newpage
To explore the effect of viewing distance, the baseline pixel height of the 3DGS avatar was measured at a viewing distance of 3.0\,m, which matches the camera distance used when capturing the training data for the avatar. At that distance, the height was 406 pixels within the image frame (520\,px $\times$ 440\,px). Using the following metric:
\[
\text{Percentage of Pixels} 
= 100 \times \frac{\text{New Pixel Height}}{\text{Baseline Pixel Height}}
\]
we generated five versions of the avatar at 100\%, 80\%, 60\%, 50\%, and 40\% of the original 3DGS avatar, each corresponding to increasing distances from the viewer (3, 3.75, 5, 6 and 7.5 metres).
 
 %The baseline viewing distance (P1 at 100\%) was set to 3.0\,m, matching the camera distance used when capturing the training data for the 3D Gaussian avatar.

Our stimuli comprised 60 video stimuli, each showing two 3DGS avatars side by side at equal distances/\%pixels, for a duration of five seconds (see Figure~\ref{fig:design}). One avatar was rendered with 202k Gaussians (considered as the pseudo ground truth or gold standard), while the other was rendered with 50k, 12k, or 3k Gaussians. For additional validation and elimination of side bias, each comparison was repeated with the gold standard and comparison avatars swapped between the left and right positions. These comparisons were presented at the five discrete viewing distances/\%pixels (100\% to 40\%).
Each video featured one of two motion types: a cyclical jogging motion or a more complex martial arts performance. These motions were sourced from the AMASS dataset~\cite{AMASS}.

\vspace{0.3cm}
\noindent\textbf{Participants and Procedure:}
A total of 35 volunteers (24M/11F, aged 18-60+) participated in the experiment. The study was conducted on the participants' own devices, including laptops, desktops, and tablets, with screen sizes ranging from less than 15 inches to more than 21 inches. However, data from two participants who used mobile devices were excluded to ensure consistency in visual presentation. We asked for participants' screen size at the beginning of the experiment, and statistical tests of the effects of screen size on the results were not significant.
% An analysis of the interaction between screen size and accuracy, shown in Figure~\ref{fig:screen_size_histogram}, revealed no significant differences in accuracy across screen sizes ranging from less than 15 inches to more than 21 inches. However, data collected from mobile devices (classified as invalid) showed a substantial bias, underscoring the necessity of their exclusion from the analysis.
% \begin{figure}[h]
%     \centering
%     \includegraphics[width=\textwidth]{figures/combined_screensize.jpg}
%     \caption{Interaction Effects of Pixels and LOD on Accuracy with Screen Size as a Factor.}
%     \label{fig:screen_size_histogram}
% \end{figure}

\begin{table}[t]
%\centering
%\vspace{0.8cm}
\caption{ANOVA showing the results of main and interaction effect tests with degrees of freedom, effect sizes ($\eta^{2}$) and ($p$) values. Significant effects are highlighted in {\color{red}red}}

\label{tab:anova_results}
\small
\begin{tabular}{llllllll}\hline\hline
\textbf{Effect Tested}  & & \multicolumn{2}{ l }{\rule{0pt}{3ex}\emph{dof}}
&\textbf{F-Test} & \textbf{$\eta^{2}$} & \textbf{p}\\\hline\hline
  Motion  & & 1 &32 & 5.92 & 0.156 & {\color{red}$<0.05$}\\
 %1	32	5.923036373	0.15618571	<0.050

  LOD  & & 2 & 64 & 65.60 & 0.672 & {\color{red}$<0.001$}\\
%2	64	65.6069869	0.672154617	<0.001

  Pixels  & & 4 & 128 & 1.15 & $<0.035$ & 0.336\\
%4	128	1.14941141	0.03467366	0.336

  Motion x LOD & & 2& 64  & 0.49 & 0.015 & 0.616\\
%2	64	0.48757764	0.015008125	0.616

  Motion x Pixels  & & 4 &128 & 0.51 & 0.016 & 0.730\\
%4	128	0.507405471	0.015608919	0.730

  LOD x Pixels  & & 5.2 &168.1 & 4.22 & 0.116 & {\color{red}$<0.001$}\\
%5.2	168.1	4.2223335	0.116567131	<0.001
 
  Motion x LOD x Pixels  & & 5.8& 187.5 & 0.47 & 0.014 & 0.825\\\hline\hline
%5.8	187.5	0.471374843	0.014516627	0.825
\end{tabular}
\end{table}

\vspace{0.3cm}
The experiment was conducted via a Qualtrics-based survey. After providing informed consent and some demographical information, participants completed a brief practice session comprising two trials to familiarize themselves with the task. The main experiment block included 60 randomized trials, evenly divided between the cyclical and complex \emph{Motion} types. In each trial, participants were presented with two animated avatars, one of which was the GS, while the other was one of the three \emph{LOD} renderings. They were both presented at the same distance, which was selected from the five levels of \emph{Pixels}. Participants were asked to identify the avatar with the higher perceived visual detail by selecting either “Left” or “Right.”

To ensure focused decision-making, the question only appeared after the video finished playing, and participants were restricted to a single viewing of the video. Once the video ended, it disappeared from the screen, preventing participants from basing their choices solely on the final image frame. This procedure was designed to minimize biases and improve the consistency of participant responses. The experiment lasted approximately 10 to 15 minutes, and participants were allowed to take breaks. For each trial, the selected avatar (Left or Right) was recorded, and accuracy was calculated based on whether they chose the side where the GS avatar was (accuracy = 1) or one of the LOD avatars (accuracy = 0). Higher accuracy therefore indicates that visual artefacts were more noticeable. These values were then averaged over the two videos with the GS on the left and on the right, to give an accuracy rate for that particular combination of variables.

% \vspace{0.3cm}
% \vspace{-10em}
\noindent\textbf{Results:}
We averaged the accuracy results over all participants, for each combination of variables, and the results can be seen in Figure ~\ref{fig:expresults}.
A three-way Analysis of Variance (ANOVA) with independent variables Motion (2) $\times$ LOD (3) $\times$ Pixels (5) and dependent variable Accuracy (i.e., the proportion of times they correctly chose the gold standard). A summary of significant and non-significant effects can be seen in Table~\ref{tab:anova_results}. The results of pairwise comparisons of differences between means (Bonferroni post-hoc tests) are shown in Table~\ref{table:posthoc}. 

\vspace{-2em}
A significant main effect of \emph{Motion ($F(1, 32) = 5.92$, $\eta^{2}=0.156$, $p <0.05$)} was found. Post-hoc analysis revealed that accuracy was higher for the complex motion (martial arts) than for the cyclical one (jogging). This indicates that visual artefacts were more obvious for the complex motion, which is the opposite of what we hypothesized in Section \ref{Intro} (Q1). 

\vspace{-2em}
We also found a significant main effect of \emph{LOD ($F(2, 64) = 65.60$, $\eta^{2}=0.672$, $p < 0.001$)}. The effect size $\eta^{2}$ of 0.67 indicates that this was quite a large effect of accuracy based on the number of Gaussians used for rendering. Post-hoc analysis revealed significant differences between all three LOD levels (all $p<0.00001$), where accuracy for LOD1 with 3k Gaussians ($m=0.93$) was higher than for both LOD2 with 12k ($m=0.81$) and LOD3 with 50k ($m=0.60$), while LOD2 accuracy was also higher than for LOD3. Consistent with our expectations (Q2), participants could successfully differentiate each LOD from the Gold Standard at above chance (i.e., 0.5) rates, in particular for the lowest quality level. 
\pagebreak

\begin{figure}[t]
    \centering    \includegraphics[width=0.9\linewidth]{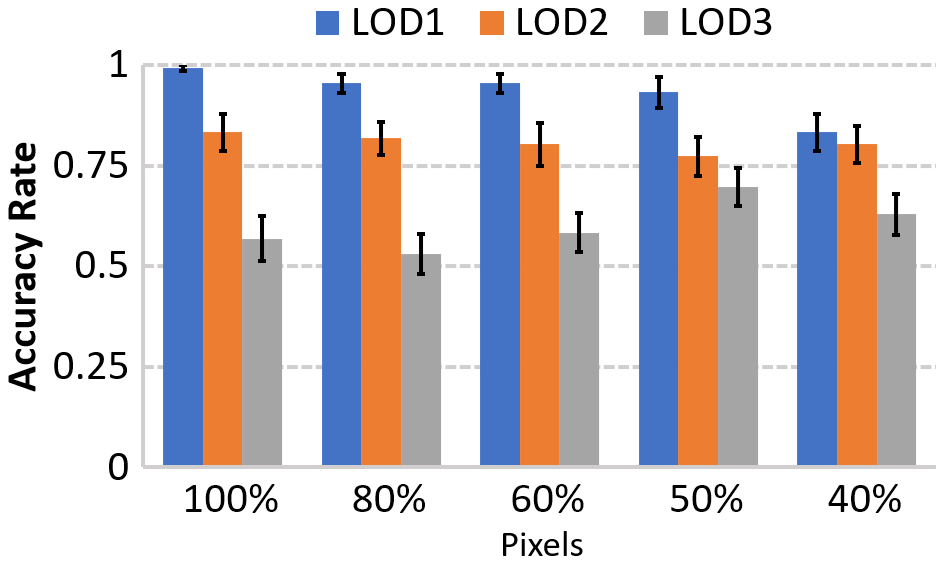}
    %\vspace{-0.2cm}
    \caption{Graph of the LOD x Pixels interaction effect}
    \vspace{0.2cm}
    \label{fig:LOD-Pixels}
\end{figure}

In negation of Q2, no significant effect of \emph{Pixels} was found, suggesting that distance/\%pixels alone does not consistently affect accuracy. However, a significant interaction effect was observed between \emph{LOD} and \emph{Pixels ($F(5.2, 168.1) = 4.22$, $\eta^{2}=0.672$, $p< 0.001$)}. This  effect can be explained by observing the graph of the interaction in Figure \ref{fig:LOD-Pixels} and the significant effects shown in Table \ref{table:posthoc}. It can be seen that, for LOD1 (with the lowest number of Gaussians), accuracy rates for avatars rendered with ($40\%$) pixels were lower than for $100\%$, $80\%$, $60\%$ and $50\%$ pixels. This indicates that increased viewing distance, corresponding to a reduced percentage of pixels, successfully masked visual artefacts for the lowest number of Gaussians, but only between the nearest and further distances. For LOD3 (with the highest number of Gaussians after the GS), accuracy for $80\%$ pixels was significantly lower than for $50\%$ pixels, although the reason for this anomaly is not obvious. 
\vspace{0.3cm}

%At closer viewing distances (P1–P2), participants exhibited higher accuracy for low-number comparisons (L1-GS), where differences were more perceptible. In contrast, accuracy declined for mid-number pairs (L2-L3) at greater distances (P4–P5), revealing a perceptual threshold beyond which number of 3D Gaussians differences became indistinguishable. 
The two-way interactions of \emph{Motion $\times$ LOD}, \emph{Motion $\times$ Pixels} and the three-way interaction of \emph{Motion $\times$ LOD $\times$ Pixels}  were not significant, which shows that that the two different motions did not change any of the other effects. Therefore, the answer to Q4 is ``yes'' for the \emph{LOD $\times$ Pixels} interaction only.

\begin{comment}
\begin{table}[h!]
    \centering
    \small
    \caption{ANOVA Results for Motion, LOD, and Pixels}
    \label{tab:anova_results}
    \begin{tabular}{lcccc}
        \toprule
        \textbf{Factor} & \textbf{F-value} & \textbf{p-value} & \textbf{Significance} \\
        \midrule
        Motion          & 9.3513   & 0.0023  & **   \\
        LOD             & 121.9137 & 0       & ***  \\
        Pixels          & 0.9904   & 0.4115  & n.s. \\
        Motion:LOD      & 0.5227   & 0.593   & n.s. \\
        Motion:Pixels   & 0.4544   & 0.7692  & n.s. \\
        LOD:Pixels      & 3.0702   & 0.0019  & **   \\
        Motion:LOD:Pixels & 0.3520 & 0.9452  & n.s. \\
        Residual        &          &         & n.s. \\
        \bottomrule
    \end{tabular}
\end{table}
\end{comment}

\begin{comment}
\paragraph{Post-Hoc Analysis.} We conducted pairwise comparisons of differences between means (Bonferroni post-hoc tests) for the significant main effects of Motion and LOD, and the interaction effect of LOD x Pixels. We found a significant difference between the two types of Motion ($p<0.05$), where accuracy rates were significantly higher for dancing ($m=0.81$) than for jogging ($m=0.75$). We also found significant differences between all three LOD levels (all $p<0.00001$), where accuracy for LOD1 with 3k Gaussians ($m=0.93$) was higher than for both LOD2 with 12k ($m=0.81$) and LOD3 with 50k ($m=0.60$), while LOD2 accuracy was also higher than for LOD3.
%as shown in Table~\ref{tab:posthoc_lod}. 

The LOD x Pixels interaction effect can be explained by observing the significant effects shown in Table \ref{table:posthoc}, and the graph of the interaction in Figure \ref{fig:LOD-Pixels}. It can be seen that, for LOD1, there were significant differences between pixels level 5 ($40\%$), where accuracy rates were lower than for levels 1 ($100\%$), 2 ($80\%$), 3 ($60\%$), and 4 ($60\%$). For LOD3, accuracy for pixel level 2 was significantly lower than for level 4.
\end{comment}

\begin{table}[t]
\centering
\caption{Significant results of Post-hoc tests (Bonferroni) for the LOD x Pixels interaction effect.}
\label{table:posthoc}
\small
\begin{tabular}{ccccc}\hline\hline
LOD &(\textbf{I}) Pixels &(\textbf{J}) Pixels &Mean diff. (\textbf{I-J}) &\textbf{p}\\ \hline\hline
1 &100\% &40\% &0.159 & {\color{red} $p<0.01$}\\
1 &80\% & 40\%&0.121 & {\color{red} $p<0.05$}\\
1  &60\% & 40\% &0.121 & {\color{red} $p<0.01$}\\
1  &50\% &40\% &0.098 & {\color{red} $p<0.05$}\\
3 &80\% &50\% &-0.167& {\color{red} $p<0.05$}\\\hline\hline
\end{tabular}
\end{table}

\begin{comment}
\begin{table}[t]
    \centering
    \small
    \caption{Post-Hoc tests of significant effects (Tukey's HSD)}
    \label{tab:posthoc}
    \begin{tabular}{lcccccc} % Adjusted to match the number of columns
        \toprule
        Group 1 & Group 2 & Mean Diff. & $p$-adj & Lower & Upper & Reject \\
        \midrule
        L1 & L2 & -0.1273 & 0.000 & -0.1778 & -0.0768 & True \\
        L1 & L3 & -0.3318 & 0.000 & -0.3823 & -0.2813 & True \\
        L2 & L3 & -0.2045 & 0.000 & -0.2550 & -0.1540 & True \\
 %       \bottomrule
 %   \end{tabular}
%\end{table}

%\begin{table}[h]
%    \centering
%    \small
%    \caption{Post-Hoc Analysis of Motion Effects Using Tukey's HSD Test}
 %   \label{tab:posthoc_motion}
 %  \begin{tabular}{lcccccc} % Adjusted to match the number of columns
%        \toprule
%        Group 1 & Group 2 & Mean Diff. & $p$-adj & Lower & Upper & Reject \\
        \midrule
        Mo1 & Mo2 & 0.0535 & 0.004 & 0.0171 & 0.0900 & True \\
        \bottomrule
    \end{tabular}
\end{table}
\end{comment}

\vspace{0.2cm}
\noindent
\textbf{Discussion:} This study provides actionable insights for optimizing LOD strategies in Gaussian-based crowd rendering. 
Inspired by the results, a progressive allocation of the number of 3D Gaussians levels is proposed to optimize computational efficiency while maintaining visual fidelity. For close distances, the gold standard (GS) with the highest number of 3D Gaussians is essential to ensure high perceptual clarity. At intermediate distances, LOD3 with the next highest \#Gaussians, provides a balance between visual quality and computational cost. For further distances, lower LOD avatars with fewer Gaussians should be sufficient to achieve further computational savings without noticeable loss in perceived quality.

%\begin{itemize}
%    \item \textbf{Motion enhances perception:} Dynamic motions (Mo2) significantly improve participants’ ability to discern number of 3D Gaussians differences, underscoring the importance of visually engaging motions in high-fidelity rendering.

%    \item \textbf{Number of 3D Gaussians and Distance Interplay:} 
%     Participants consistently distinguished lowest-number pairs (L1-GS) across all viewing distances. Their perceptual clarity was highest at close distances (P1–P2), where finer details of the gold standard (GS) number of 3D Gaussians were more apparent. As the viewing distance increased (P3–P5), mid-number pairs (L2-L3) became less distinguishable, indicating that lower-number levels can achieve comparable perceptual outcomes at these distances with significant computational savings.

\section{Conclusions and Future Work}
This study systematically evaluates the perceived LOD in CrowdSplat, focusing on how the number of 3D Gaussians, motion, and viewing distance influence visual perception. The results demonstrate the potential of this approach for optimizing rendering strategies to achieve computational efficiency without sacrificing visual fidelity. 
However, several \emph{limitations} provide opportunities for future exploration:
\vspace{-0.2cm}

\begin{itemize}
\item 3DGS uses multiple-view images as input to reconstruct scenes. The opacity $\alpha$ of 3D Gaussians is optimized between 0 and 1. However, GaussianAvatar \cite{GSA} fixes the opacity at $\alpha = 1$ to keep all 3D Gaussians visible, forcing the network to predict accurate 3D Gaussian positions. When multiple opaque 3D Gaussians are projected onto a single pixel, artefacts may appear due to the overlapping of 3D Gaussians at greater distances. 
\vspace{-0.2cm}
\item  Unintended visual artefacts, such as white spots around the hands, dotted lines, or text on the avatars' clothing, were noted by participants as influencing their judgments, potentially introducing bias into the results. Addressing these artefacts in future experiments will help eliminate such confounding factors. 
\vspace{-0.2cm}
\item The online experimental environment relied on participants using personal devices with varying display size and quality. This inconsistency may have affected their ability to perceive fine-grained details, particularly at greater distances. Future studies are needed where  viewing conditions are standardized to ensure consistency across participants. 
\vspace{-0.2cm}
\item While the study examined motion, the number of 3D Gaussians, and distance, it did not account for other perceptual factors, such as lighting conditions
or environmental complexity. Incorporating these variables in future research would enhance the applicability and robustness of the proposed thresholds.
\end{itemize}
\vspace{-0.2cm}

%The results indicate the importance of visually engaging motions and the strategic resource allocation for LOD to align with perceptual clarity across varying viewing distances.
In conclusion, despite these limitations, our results provide a solid foundation for the improvement of Gaussian-based crowd rendering systems, with practical implications for real-time applications requiring scalable and realistic visualizations.

\acknowledgments{
This work was conducted with the financial support of the Research Ireland Centre for Research Training in Digitally-Enhanced Reality (d-real) under Grant No. 18/CRT/6224. For the purpose of Open Access, the author has applied a CC BY public copyright licence to any Author Accepted Manuscript version arising from this submission.}

\bibliographystyle{abbrv-doi}

\bibliography{template}

\begin{thebibliography}{10}

\bibitem{I1}
A.~Aubel, R.~Boulic, and D.~Thalmann.
\newblock Animated impostors for real-time display of numerous virtual humans.
\newblock In {\em International Conference on Virtual Worlds}, pp. 14--28. Springer, 1998.

\bibitem{Bar05}
J.~A. B{\ae}rentzen.
\newblock Hardware-accelerated point generation and rendering of point-based impostors.
\newblock {\em Journal of Graphics Tools}, 10(2):1--12, 2005.

\bibitem{Beacco12}
A.~Beacco, C.~And{\'u}jar, N.~Pelechano, and B.~Spanlang.
\newblock Efficient rendering of animated characters through optimized per-joint impostors.
\newblock {\em Computer Animation and Virtual Worlds}, 23(1):33--47, 2012.

\bibitem{Beacco11}
A.~Beacco, B.~Spanlang, C.~Andujar, and N.~Pelechano.
\newblock A flexible approach for output-sensitive rendering of animated characters.
\newblock In {\em Computer Graphics Forum}, vol.~30, pp. 2328--2340. Wiley Online Library, 2011.

\bibitem{I3}
S.~Dobbyn, J.~Hamill, K.~O'Conor, and C.~O'Sullivan.
\newblock Geopostors: a real-time geometry/impostor crowd rendering system.
\newblock In {\em Proceedings of the 2005 symposium on Interactive 3D graphics and games}, pp. 95--102, 2005.

\bibitem{RW4}
X.~Dongye, H.~Guo, L.~Luo, H.~Jiang, Y.~Bao, Z.~Tian, and D.~Weng.
\newblock Lodavatar: Hierarchical embedding and adaptive levels of detail with gaussian splatting for enhanced human avatars.
\newblock {\em arXiv preprint arXiv:2410.20789}, 2024.

\bibitem{GSA}
L.~Hu, H.~Zhang, Y.~Zhang, B.~Zhou, B.~Liu, S.~Zhang, and L.~Nie.
\newblock Gaussianavatar: Towards realistic human avatar modeling from a single video via animatable 3d gaussians.
\newblock In {\em Proceedings of the IEEE/CVF Conference on Computer Vision and Pattern Recognition}, pp. 634--644, 2024.

\bibitem{hu2023gauhuman}
S.~Hu and Z.~Liu.
\newblock Gauhuman: Articulated gaussian splatting from monocular human videos.
\newblock {\em arXiv preprint arXiv:}, 2023.

\bibitem{Jiang2022InstantAvatarLA}
T.~Jiang, X.~Chen, J.~Song, and O.~Hilliges.
\newblock Instantavatar: Learning avatars from monocular video in 60 seconds.
\newblock {\em 2023 IEEE/CVF Conference on Computer Vision and Pattern Recognition (CVPR)}, pp. 16922--16932, 2022.

\bibitem{PP}
L.~Kavan, S.~Dobbyn, S.~Collins, J.~{\v{Z}}{\'a}ra, and C.~O'Sullivan.
\newblock Polypostors: 2d polygonal impostors for 3d crowds.
\newblock In {\em Proceedings of the 2008 symposium on Interactive 3D graphics and games}, pp. 149--155, 2008.

\bibitem{3dgs}
B.~Kerbl, G.~Kopanas, T.~Leimk{\"u}hler, and G.~Drettakis.
\newblock 3d gaussian splatting for real-time radiance field rendering.
\newblock {\em ACM Trans. Graph.}, 42(4):139--1, 2023.

\bibitem{hierarchicalgaussians24}
B.~Kerbl, A.~Meuleman, G.~Kopanas, M.~Wimmer, A.~Lanvin, and G.~Drettakis.
\newblock A hierarchical 3d gaussian representation for real-time rendering of very large datasets.
\newblock {\em ACM Transactions on Graphics}, 43(4), July 2024.

\bibitem{hugs}
M.~Kocabas, R.~Chang, J.~Gabriel, O.~Tuzel, and A.~Ranjan.
\newblock Hugs: Human gaussian splats, 2023.

\bibitem{LS09}
E.~Landreneau and S.~Schaefer.
\newblock Simplification of articulated meshes.
\newblock In {\em Computer Graphics Forum}, vol.~28, pp. 347--353. Wiley Online Library, 2009.

\bibitem{Lemonari:2022:Authoring}
M.~Lemonari, R.~Blanco, P.~Charalambous, N.~Pelechano, M.~Avraamides, J.~Pettré, and Y.~Chrysanthou.
\newblock {Authoring Virtual Crowds: A Survey}.
\newblock {\em Computer Graphics Forum}, 2022. doi: {{%
10\hspace{.1pt}\discretionary{.}{%
}{.}\hspace{.4pt}1111\discretionary{/}{%
}{/}cgf\hspace{.1pt}\discretionary{.}{%
}{.}\hspace{.4pt}14506}}


\bibitem{liu2024citygaussian}
Y.~Liu, H.~Guan, C.~Luo, L.~Fan, J.~Peng, and Z.~Zhang.
\newblock Citygaussian: Real-time high-quality large-scale scene rendering with gaussians, 2024.

\bibitem{AMASS}
N.~Mahmood, N.~Ghorbani, N.~F. Troje, G.~Pons-Moll, and M.~J. Black.
\newblock Amass: Archive of motion capture as surface shapes.
\newblock In {\em Proceedings of the IEEE/CVF international conference on computer vision}, pp. 5442--5451, 2019.

\bibitem{RW3}
R.~McDonnell, S.~Dobbyn, S.~Collins, and C.~O’Sullivan.
\newblock Perceptual evaluation of lod clothing for virtual humans.
\newblock In {\em Symposium on Computer Animation}, pp. 117--126. Citeseer, 2006.

\bibitem{mcdonnell2008clone}
R.~McDonnell, M.~Larkin, S.~Dobbyn, S.~Collins, and C.~O'Sullivan.
\newblock Clone attack! perception of crowd variety.
\newblock In {\em ACM SIGGRAPH 2008 papers}, pp. 1--8. ACM, 2008.

\bibitem{RW2}
R.~McDonnell, C.~O'Sullivan, and S.~Dobbyn.
\newblock Lod human representations: A comparative study.
\newblock In {\em First International Conference on Crowds Simulation (V-CROWD'05)}, pp. 101--115, 2005.

\bibitem{qian20233dgsavatar}
Z.~Qian, S.~Wang, M.~Mihajlovic, A.~Geiger, and S.~Tang.
\newblock 3dgs-avatar: Animatable avatars via deformable 3d gaussian splatting.
\newblock In {\em CVPR}, 2024.

\bibitem{CrowdSplat}
X.~Sun, Y.~Xu, J.~Dingliana, and C.~O’Sullivan.
\newblock Crowdsplat: exploring gaussian splatting for crowd rendering.
\newblock In {\em IET Conference Proceedings CP887}, vol. 2024, pp. 311--314. IET, 2024.

\bibitem{TBP08}
N.~Tatarchuk, J.~Barczak, and B.~Purnomo.
\newblock Gpu tessellation for detailed, animated crowds.
\newblock {\em SIGGRAPH Asia}, 2008.

\bibitem{TLC2002}
F.~Tecchia, C.~Loscos, and Y.~Chrysanthou.
\newblock Image-based crowd rendering.
\newblock {\em IEEE computer graphics and applications}, 22(2):36--43, 2002.

\bibitem{WS02}
M.~Wand and W.~Stra{\ss}er.
\newblock Multi-resolution rendering of complex animated scenes.
\newblock In {\em Computer Graphics Forum}, vol.~21, pp. 483--491. Wiley Online Library, 2002.

\bibitem{Weng2022HumanNeRFFR}
C.-Y. Weng, B.~Curless, P.~P. Srinivasan, J.~T. Barron, and I.~Kemelmacher-Shlizerman.
\newblock Humannerf: Free-viewpoint rendering of moving people from monocular video.
\newblock {\em 2022 IEEE/CVF Conference on Computer Vision and Pattern Recognition (CVPR)}, pp. 16189--16199, 2022.

\bibitem{RW1}
F.~Yang, J.~Shabo, A.~Qureshi, and C.~Peters.
\newblock Do you see groups? the impact of crowd density and viewpoint on the perception of groups.
\newblock In {\em Proceedings of the 18th International Conference on Intelligent Virtual Agents}, pp. 313--318, 2018.

\bibitem{zielonka25dega}
W.~Zielonka, T.~Bagautdinov, S.~Saito, M.~Zollhöfer, J.~Thies, and J.~Romero.
\newblock Drivable 3d gaussian avatars.
\newblock In {\em International Conference on 3D Vision (3DV)}, March 2025.

\end{thebibliography}
\end{document}